\documentclass[letterpaper, 10 pt, conference]{ieeeconf}  

\IEEEoverridecommandlockouts                              

\usepackage{graphics} 
\usepackage{epsfig} 
\usepackage{mathptmx} 
\usepackage{graphicx}
\usepackage{booktabs}
\usepackage{adjustbox}
\usepackage{todonotes}
\usepackage{caption}
\usepackage{soul}
\graphicspath{{figures/}}
\usepackage{amsmath}

\newcommand\atwnote[1]{}
\newcommand\wwnote[1]{}

\newcommand\wwrevise[1]{\textcolor{black}{#1}}

\title{\LARGE \bf
Part Segmentation for Highly Accurate Deformable Tracking in Occlusions via Fully Convolutional Neural Networks
}

\author{Weilin Wan$^{1}$, Aaron Walsman$^{1}$, and Dieter Fox$^{1,2}$
\thanks{$^{1}$Paul G. Allen School of Computer Science and Engineering, University of Washington.}%
\thanks{$^{2}$NVIDIA Research Seattle.}%
}

\begin{document}

\maketitle
\thispagestyle{empty}
\pagestyle{empty}



\begin{abstract}
Successfully tracking the human body is an important perceptual challenge for robots that must work around people.
Existing methods fall into two broad categories: geometric tracking and direct pose estimation using machine learning.  While recent work has shown direct estimation techniques can be quite powerful, geometric tracking methods using point clouds can provide a very high level of 3D accuracy which is necessary for many robotic applications.
However these approaches can have difficulty in clutter when large portions of the subject are occluded.
To overcome this limitation, we propose
a solution based on fully convolutional neural networks (FCN).
We develop an optimized Fast-FCN network architecture for our application which
allows us to filter observed point clouds and improve tracking accuracy while maintaining
interactive frame rates.
We also show that this model can be trained with a limited number
of examples and almost no manual labelling by using an existing geometric tracker and data augmentation to automatically generate segmentation maps.
We demonstrate the accuracy of our full system by comparing it against an existing geometric tracker, and show significant improvement in these challenging scenarios.

\end{abstract}

\section{Introduction}
Human pose tracking in 3D 
is required 
for many robotic applications, including robotic medical care and personal assistance.
Recent work 
in pose estimation and human body tracking 
has achieved accurate results in 
uncluttered scenarios, 
especially when depth information is available \cite{schmidt2014dart, ye2014real, walsman2017dynamic}.
However, in most practical applications, humans often interact closely with objects
and are often partially occluded from the
 view of the camera\atwnote{\st{s}}. In this case, the RGB-D information generated by the objects 
 can interfere with the geometric computation and cause errors in tracking.\par 
\begin{figure}[t]
  \centering
  \hbox{\hspace{-0.5em}\includegraphics[scale=0.38]{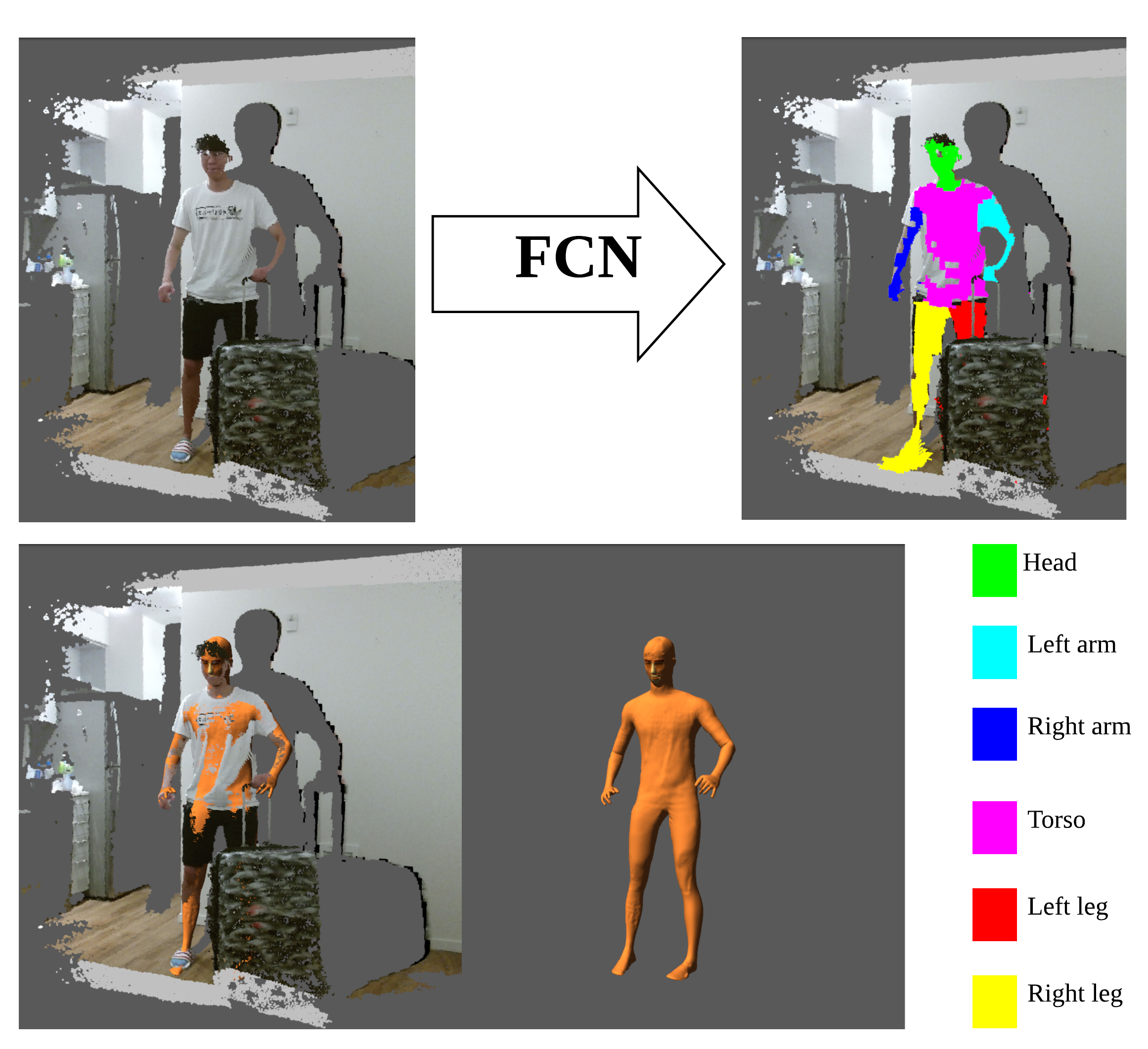}}
  \caption{Our system tracking a frame with occlusion. The top left is the observation point cloud. The top right is the semantic segmentation maps generated by our FCN network. The bottom shows the result of tracking.}
  \label{fig:one}
\end{figure}
\atwnote{\st{Thus,}}In this paper, we focus on improving the robustness and reliability 
of pose tracking with occlusions 
using RGBD data. In particular, we propose a method based on FCN, which is developed from \atwnote{\st{regular}} deep convolutional neural networks by adding deconvolutional layers, allowing the network to output pixel-wise classifications. Rather than individual labels, the output of FCNs can be the same resolution as the input images.
We propose 
a lightweight Fast-FCN, 
and show that it can be incorporated into a geometric model-based tracker to drastically reduce errors caused by clutter and occlusion and to initialize the model pose.  Figure \ref{fig:one} illustrates the structure of our system.
\atwnote{\st{Actually,}} FCNs have been applied by researchers in semantic segmentation analysis and pose estimation with RGB data, but we focus \atwnote{\st{ing}}on 
optimization-based 3D tracking in this paper, 
in order to achieve the extremely high accuracy required in many practical robotics applications. 
Our model can achieve \atwnote{\st{an}}accurate pixel-wise predictions with a simplified network structure. The Fast-FCN architecture that we propose\atwnote{\st{d}} is able to achieve a much better run time performance with a minor accuracy cost in our task, which means we can run it along with geometric optimization and keep the overall system running at interactive frame rates.

Fully Convolutional Networks often require a large amount of training data which can be time-consuming and expensive to label by hand.  In order to address this, we built a new RGBD dataset by using an existing geometric tracker 
to label human poses in unoccluded scenarios.  We then randomly added artificial occluding objects to these videos and trained our network to label human body parts in the presence of occlusion.  This data augmentation technique was extremely effective and allowed us to train these networks with almost no manual labelling.  The geometric tracker alone frequently fails on this augmented dataset, but our new tracker that incorporates the Fast-FCN model is able to remove these occlusions and track the human poses successfully, both in this augmented dataset and in naturally generated sequences with occlusion.
\atwnote{Let's rethink the paragraph above slightly.  I think we can emphasize why we did all of this if we say something like: "Fully Convolutional Networks often require a large amount of training data which can be time-consuming and expensive to label by hand.  In order to address this, we built a new RGBD dataset by using an existing geometric tracker \cite{walsman2017dynamic} to label human poses in unoccluded scenarios.  We then randomly added artificial occluding objects to and these videos and trained our network to label human body parts in the presence of occlusion.  This data augmentation technique was extremely effective and allowed us to train these networks with almost no manual labelling.  The geometric tracker alone frequently fails on this augmented dataset, but our new tracker that incorporates the Fast-FCN model is able to remove these occlusions and track the human poses successfully, both in this agumented dataset and in naturally generated sequences with occlusion."}\par
The paper is organized as follows. In Section II, we discuss the related work\atwnote{\st{s}}. Section III describes the methodology of our tracking system. Section IV introduces the dataset we 
used for this task. The experiments and evaluations are documented in Section V.
Section VI 
concludes.

\section{Related Work}

Prior work in human pose estimation can be divided into two basic
approaches: geometric tracking and discriminative prediction using
machine learning.
Geometric tracking had many early successes \cite{
grest2005nonlinear, ganapathi2012real, schmidt2014dart}.
The core idea is to optimize the pose of a kinematic human model by reducing
an error function designed to describe the distance between the current pose and an observation.  In recent years this has been extended to deformable models \cite{ye2014real, walsman2017dynamic}.  The downside of
these approaches is that they often require some pose initialization, and
can be sensitive to occlusion and interference from nearby objects.  In
our approach, we avoid these problems by incorporating a fast and effective
discriminative model to initialize the pose when the model is out of place
and filter points that do not belong to the body.

Discriminative methods also have a rich history \cite{
haque2016towards, mehta2017vnect}.  In these approaches, the
position of joint locations are predicted directly from observations.
The skeleton tracking in the ground-breaking Microsoft Kinect used this
approach to estimate the pose and gestures of humans playing video games
\cite{shotton2011real}.  While the Kinect introduced cheap depth-sensing
to the wider robotics community, there have also been several attempts
to estimate 2D and 3D human pose directly from RGB images without the use of
depth information \cite{ferrari2008progressive,
iqbal2016posetrack, carreira2016human}.
More recently, discriminative methods using deep learning
have shown remarkable success in this domain \cite{cao2016realtime,
papandreou2017towards, mehta2018single}.
Recent methods have also been able to generate fine-grained part
locations\cite{guler2018densepose} and even 3D reconstructions of human pose \cite{varol2018bodynet, omran2018neural} from RGB images, although these methods have not been shown
to produce the 3D accuracy necessary for safe physical human-robot interaction that is available using commercial depth sensors.
In contrast, our approach uses an efficient discriminative model in 
conjunction with a 3D pose tracker designed to produce highly accurate
spatial information.  This plays to the strengths of both approaches: the 
descriminative component is able to locate subject anywhere in the frame
and make important semantic decisions about which points belong to the body 
and which do not, which enables the tracking component to produce highly 
accurate 3D pose information even in the presence of clutter and occlusion.

There have also been many hybrid approaches to this problem \cite{
ganapathi2010real, helten2013personalization, tompson2014real}.
A common approach here is to use a discriminative component to construct 
one or more additional loss terms for the tracker's optimization to drive the pose toward detected body parts.  In our approach, we use a fully convolutional network to
segment the observed image into regions corresponding to various body parts
and the background.  We then use this segmentation to inform the association
between our model and the point-cloud.
We also augment our loss function to drive
joints toward detected regions, but this
is only used when the tracker is being initialized or has lost track of the subject.

Fully convolutional networks (FCN) have become quite popular for semantic
segmentation \cite{noh2015learning, long2015fully, ronneberger2015u}.
These methods produce pixel-level predictions by first downsampling an
image using convolutional layers to predict semantic features
and then upsampling these features using deconvolutional layers to
generate a high-resolution output.  These approaches have been successfully
applied to 2D human body part segmentation and pose estimation
in many recent works \cite{xia2017joint, oliveira2016deep, bulat2016human}.
While these techniques are able to attain high accuracy in terms of 2D
joint locations and/or high IoU for part segmentation, we found that
when used in conjunction with a good articulated tracker these perfomance
metrics were not strongly indicative of final 3D tracking performance.
We therefore opted to use a lightweight and fast network that allowed
us to maintain both high accuracy and frame rate.

It is also worth noting that discriminative methods typically require
very large datasets to train effectively.  Here we again use the synergy
between our two components to side-step this issue.  While our learned
model also requires significant training data, we were able to
generate this easily by using our tracker to label sequences, and then
augmenting these sequences with artificial occlusions and distractor
objects.  This allowed us to train our model in these difficult scenarios
without resorting to manual data labelling.

\begin{figure*}[thpb]

 \center

  \includegraphics[width=\textwidth]{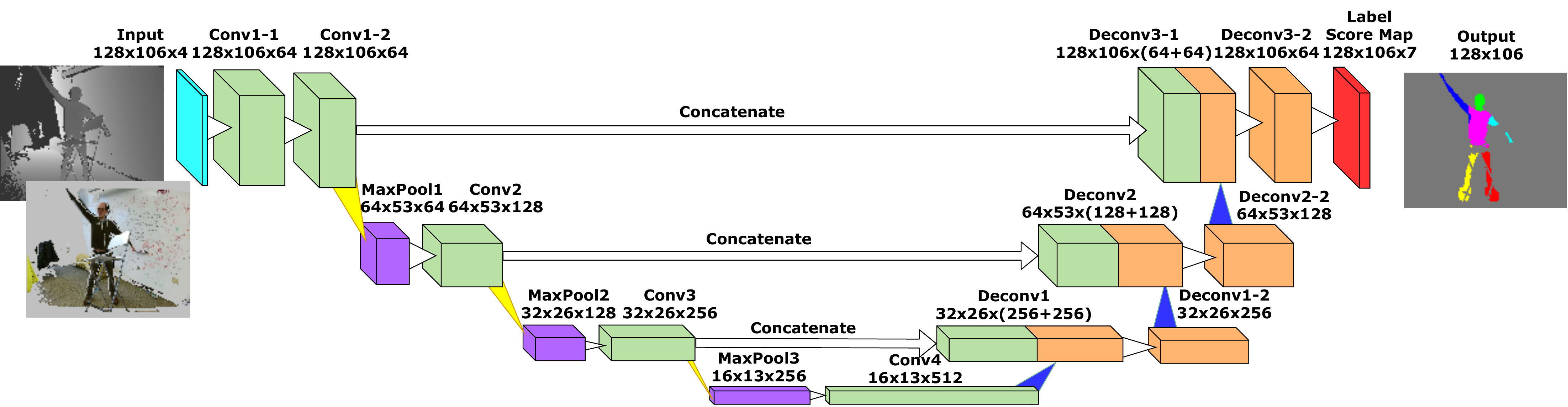}

  \caption{The architecture of our Fast-FCN model, predicting a body semantic segmentation map from a RGB-D video frame input. Convolutional layers, max pool layers and deconvolutional layers are denoted as green, purple, and orange, respectively. White triangles denote $3\times3$ convolution, yellow triangles denote $2\times2$ max pooling, and blue triangles denote $4\times4$ deconvolution.
  }

  \label{fig:network_diagram}

\end{figure*}

\section{Methodology}
Our system consists of an FCN based classifier and an optimization-based tracker similar to \cite{walsman2017dynamic}.  The Fast-FCN model we propose 
is able to provide pixel-wise semantic segmentation of body parts. \atwnote{In this paragraph, I would talk about tracking first since it's our most important component.  This means I would move the sentence above after the sentence below, and simplify the two sentences about initialization and joint error penalty into one.}When tracking, instead of computing data association to the entire point cloud, we restrict ourselves to associations over the points labelled by the FCN.
We also use these segmentation masks to
provide a coarse error signal for the pose
when initializing the model and when the
optimizer loses track of the subject.
\atwnote{You could also remove these last two sentences to save a little space.}
Section \ref{sec:architecture} explains the architecture of our FCN. Section \ref{sec:optimization}
introduces how we use the semantic segmentation in pose optimization and Section \ref{sec:targets} discusses how the
error signal for the pose is computed
when initializing the model and recovering
from tracking failures.

\subsection{Fast-FCN Architecture}
\label{sec:architecture}
In this 
section, we focus on producing pixel-wise semantic segmentation of body parts for pose initialization and optimization via a Fully Convolutional Network (FCN).
We use a second generation Microsoft Kinect that produces RGBD data with a resolution of $512\times424$. 
For performance reasons, we re-scale the input data to be $128\times106$ with four channels (RGB-D). The architecture of our network is illustrated in Figure \ref{fig:network_diagram}. The first half of network architecture down-samples the input data from $128 \times 106 \times 4$ to $16 \times 13 \times 512$. In each convolutional layer, we use the rectified linear unit (ReLU) \atwnote{\st{as}} activation function. The second half up-samples the data symmetrically and produces a semantic logistic output for each pixel. In order to keep the system running in real-time,
we must balance accuracy and complexity. 
This tradeoff is discussed further in Section \ref{sec:experiments}. \par
When estimating segmentations we consider a label set Y = \{background, head, torso, right arm, left arm, right leg, left leg\}.
The output is a single channel potential map with the same size as the input image, where $p_{iy} \in Y$ denotes the label of the $i^{th}$ pixel in the image.
In our application, the most important task is to filter out\atwnote{\st{the interfered}} objects
that may interfere with the tracking by marking them as background. Inspired by \cite{shen2015deepcontour}, in which a modification of standard cross-entropy loss is proposed for imbalanced labels,  we define our loss $L$  on frame $f$ by using cross-entropy 
along with a class imbalance modifier for all the pixels in $f$. First, for the $i^{th}$ pixel in $f$, the cross-entropy is calculated as:
\[
H_{i}=-\sum\limits_{y=1}^{|Y|} p_{iy} \cdot \textrm{log}(p'_{iy})
\]
where $p'_{iy}$ is the probability of pixel $i$ labeled as $y$, which is calculated using the softmax function given the network output label score $\{x_{iy}\}_{y=1}^{|Y|}$:
\[
p'_{iy}=\frac{\textrm{exp}(x_{iy})}{\sum_{y'=1}^{|Y|}\textrm{exp}(x_{iy'})}
\]
We add a class imbalance
modifier M to address the difference between background and body parts. At pixel $i$, M is defined as:
\[
    M_{i}=
    \begin{cases}
      -\textrm{log}(p'_{iy=\textrm{background}}), & \text{if}\ p_{iy} \textrm{ is background} \\
      -\textrm{log}(1-p'_{iy=\textrm{background}}), & \text{otherwise}
    \end{cases}
\]
Hence, for the frame $f$ with $n_{p}$ pixels in total, our loss $L$ is calculated as:
\[
L = \frac{1}{n_{p}}\sum\limits_{i=1}^{n_{p}}(H_{i}+\lambda M_{i})
\]
where $\lambda$ is a scalar factor for adjusting the effect of M.
The Fast-FCN architecture was implemented in the Tensorflow \cite{abadi2016tensorflow} machine learning framework
using the Adam optimization algorithm.

\subsection{Optimization with Semantic Filtering}
\label{sec:optimization}
To achieve high accuracy in tracking, we combined our Fast-FCN with the 3D articulated human model based generative tracking method proposed in
\cite{walsman2017dynamic}.
We modified the process of calculating the offsets between the model and the observations.


Our approach follows the geometric tracking paradigm
in which we have an articulated kinematic model that
we fit to point cloud observations.
Each vertex $v$ on the model is attached to a kinematic
skeleton via smooth-skinning \cite{kavan2008geometric}
and has a semantic label $v_y \in Y$
Each step of the tracker consists of two phases.
In the data association phase, a residual
is computed based on associations between model vertices and nearby
point cloud observations.  Once we have this residual, the
derivative of the model pose with respect to the combined
error is computed and an optimization step is taken to
minimize this error.  There are many options and techniques
for data association.  Schmidt et al. \cite{schmidt2014dart}
use a model made up of signed distance fields in order to
quickly compute distances between the point cloud
observations and model links, associating each point to
the closest link.  Ye and Yang \cite{ye2014real} use
a Gaussian Mixture Model of a random sample of their
observations and model vertices.  Walsman et. al
\cite{walsman2017dynamic} use a window-based
search technique to match model vertices with the
point cloud.  Despite this variety, all of these
techniques have one thing in common: they use physical
distance to compute association.

In contrast, our
technique first uses the Fast-FCN model described above
to construct semantic labels for each pixel.  Once these
labels have been computed, we use the window-based
search from \cite{walsman2017dynamic}, but only allow
associations between vertices and observations that have
the same label.  This prevents two common failure modes.
First, observation points that are not part of the subject
are labelled background, which prevents occluding objects
and clutter from attracting the model away from its correct
pose.  Second, this prevents different body parts from
interfering with each other and helps the tracker resolve
ambiguity when they get too close to each other.
Once these associations have been made, the rest of
our tracking procedure follows \cite{walsman2017dynamic}
with some minor modifications described below.


\subsection{Initialization and Tracking Recovery}
\label{sec:targets}

\atwnote{It seems like this section could be simplified/compressed somewhat to save space.}

One issue common to many model-based trackers is the requirement of a good initialization on each frame.  This means that if the model
ever strays too far from the subject's actual pose, the tracker may not be able to follow and could get stuck in a bad local optimum.
This is especially problematic at the beginning of a sequence where the subject's initial position is completely unknown.
Many discriminative and hybrid techniques such as \cite{ganapathi2010real, shotton2011real} attempt to remedy this by
detecting body part positions directly from the observations.  In hybrid model-based approaches the distance between the model joint locations and these
detected positions can then be used to construct an additional term in the optimizer's loss function.  We take a similar approach and use our
body-part segmentation to generate approximate centroids for each body part.  In practice, we found that these detected centroids
do not need to be highly accurate, and that if they are able to drive the model to within 20cm
of the true position, the vertex-based tracking mechanism is able to take over and accurately recover the pose.

\begin{figure}[t]
  \centering
  \includegraphics[scale=0.35]{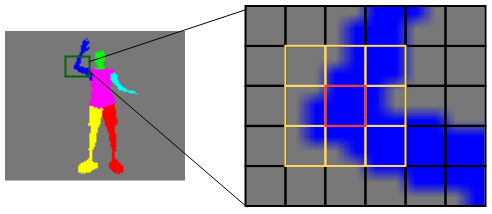}
  \caption{
  The red square here has been selected as
  the target for the right arm due to the
  density of right arm pixels in the yellow
  area.
  }
  \label{fig:targetsGen}
\end{figure}

\atwnote{After looking at the following paragraph, the technique here seems a little strange.
I understand why we don't just use the mean position, but doesn't this mean that you will just
find the thickest part of each body part?  You said you are using the elbows/knees/middle of the torso/head joints?
Isn't this going to pull the knees up to the hips?  Also, I think we should shorten this paragraph.}

To generate centroids, we calculate one target position
$t_{y} \in \rm I\!R^{3}$ for each body part label
in $Y$ except the background based on
the semantic segmentation map.  Rather than
use an expensive detection mechanism to
accurately locate joint positions, we use
a binning approach to pick
the point with the highest approximate
density as shown in Figure \ref{fig:targetsGen}.  We then add a loss to the
tracking error term of the form:
\[
\sum_{y \in Y} ||t_{y} - p_{y}||_2^2
\]

where $p_y$ is the position of a joint
in the model that corresponds to the
center of the body part (the elbow joints
in the arms, knee joints in the legs,
chest joint in the torso, and base of the
skull for the head).  Note that while
the point with the highest density does
not always correspond to these features,
this method is only used to get the model
into a position close enough for the
3D tracker to take over.  With that in
mind, we set a low weight on this error term
so that it does not interfere
with the tracker's normal operation except
during initialization and tracking failure.


\section{Data generation}
For training and evaluation, we made an RGB-D indoor human video dataset by recording eight individuals performing a variety of motions.
We applied the tracking system \mbox{in \cite{walsman2017dynamic}} to
these videos and used the output to generate ground truth for each video frame. We then augmented the dataset by inserting RGB-D images of different objects into each video sequence as clutter and occlusion. This approach allowed us to simultaneously generate ground truth information without expensive manual labelling while also producing difficult scenarios that the tracker could not handle alone.

The video sequences were made using a second generation Microsoft Kinect \cite{sell2014xbox}.
Each video contains 300 frames and the resolution of the depth data is $512\times424$.
There are thirty-two video sequences of human body
motion without occlusions, as well as seven video sequences of human interacting with everyday objects including a table, chair, suitcase and guitar.
The thirty-two sequences without occlusions form the raw material for our
augmented data, while the remaining seven sequences are used as an additional test set to ensure that our
trained model works on natural data.

\wwrevise{In our dataset, we mainly focus on natural poses, such as standing with arms swinging, walking around pulling a suitcase and sitting on stools. Since the goal of Fast-FCN classifier is to help the internal generative tracking method to maintain its tracking accuracy in occlusions, we only include poses that internal tracking methods are
capable of recovering when no occlusions are present.}

\subsection{Generating Labels}
In order to generate labels for the augmented data, we first generated a pixel-wise ground truth body part map for each
frame in the unoccluded sequences.
The ground-truth maps are
$128\times106$ pixels and have seven labels: background, head, left arm, torso, right arm, left leg, right leg.
As we mentioned above, we applied a model based tracking system
to each sequence.
We assigned each vertex of the model one of the corresponding labels, and then during tracking we used the geometric
data association computed by tracker to transfer these labels to the point cloud.
After running each sequence, we manually discarded any frames where the tracker failed, which was the only manual process
in this procedure.
\subsection{Data Augmentation
}
To simulate the inferences caused by occluding objects and clutter, we deliberately inserted different objects into the non-occluded videos.
We cropped both RGB and depth images of different objects, including a sofa, lamp, bedside table, and bench from the SUN v2 RGB-D dataset \cite{song2015sun}.  Also, we collected our own RGB-D object images, including a suitcase, desk-chair, guitar, paper box, and stool using the Kinect. We augmented the dataset by overlaying these different objects onto each video sequence with different scaling, rotation, and depth offsets. \wwrevise{In general we control the objects to occlude around one-third to a half of the person.}

\begin{figure}[h]
  \centering
  \hbox{
  \includegraphics[scale=0.33]{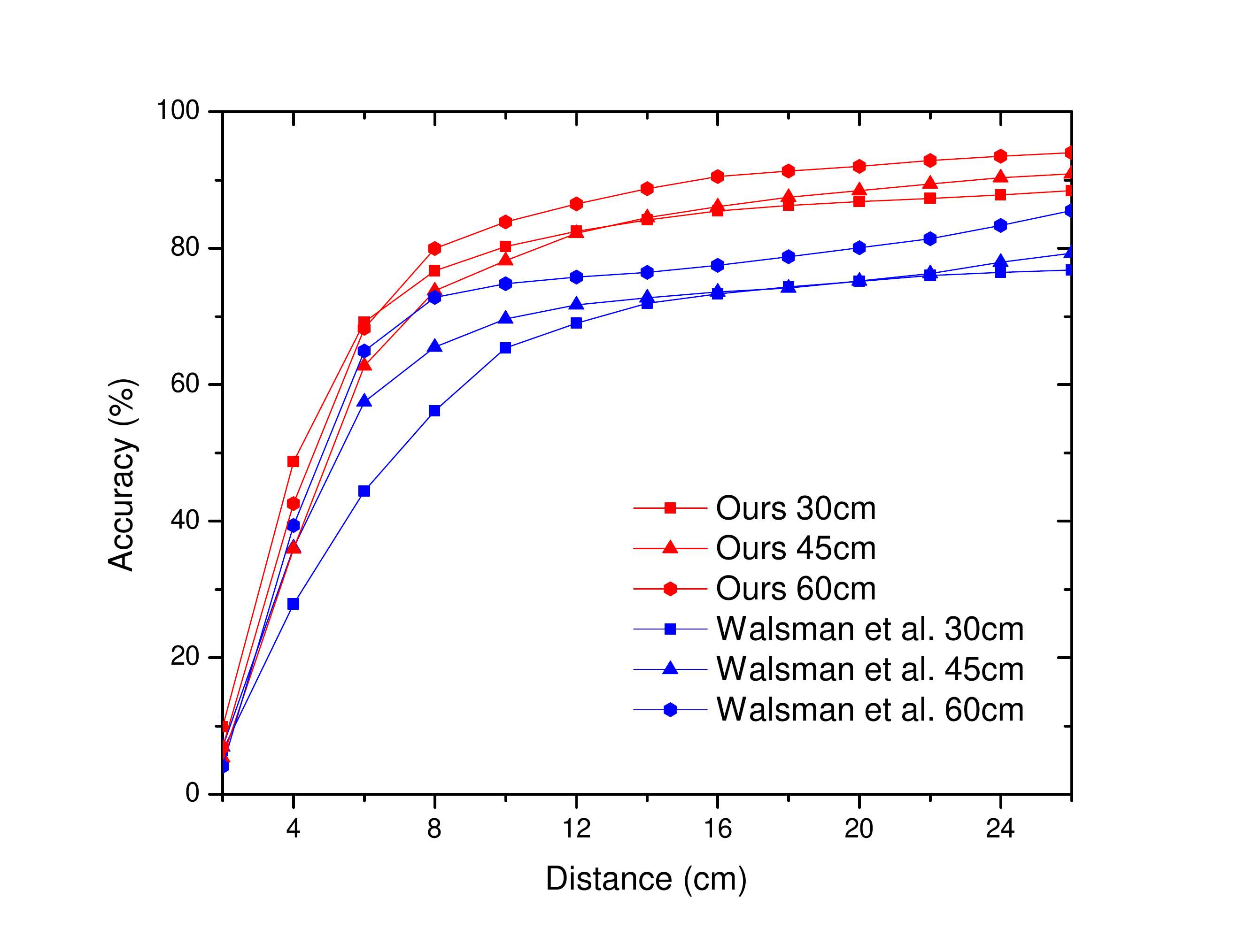}}
  \caption{A comparison of tracking performance on the augmented object-insertion data compared against \cite{walsman2017dynamic}. The x axis represents distance thresholds from the ground truth joint positions. Each curve represents the percentage of joint positions (y-axis) that are within the distance indicated from the
  ground truth locations (x-axis).
  }
  \label{fig:accuracy_plot}
\end{figure}
\section{Experiments}
\label{sec:experiments}
\begin{figure*}[thpb]
  \centering
  \hbox{\hspace{-0.8em}\includegraphics[scale=0.338]{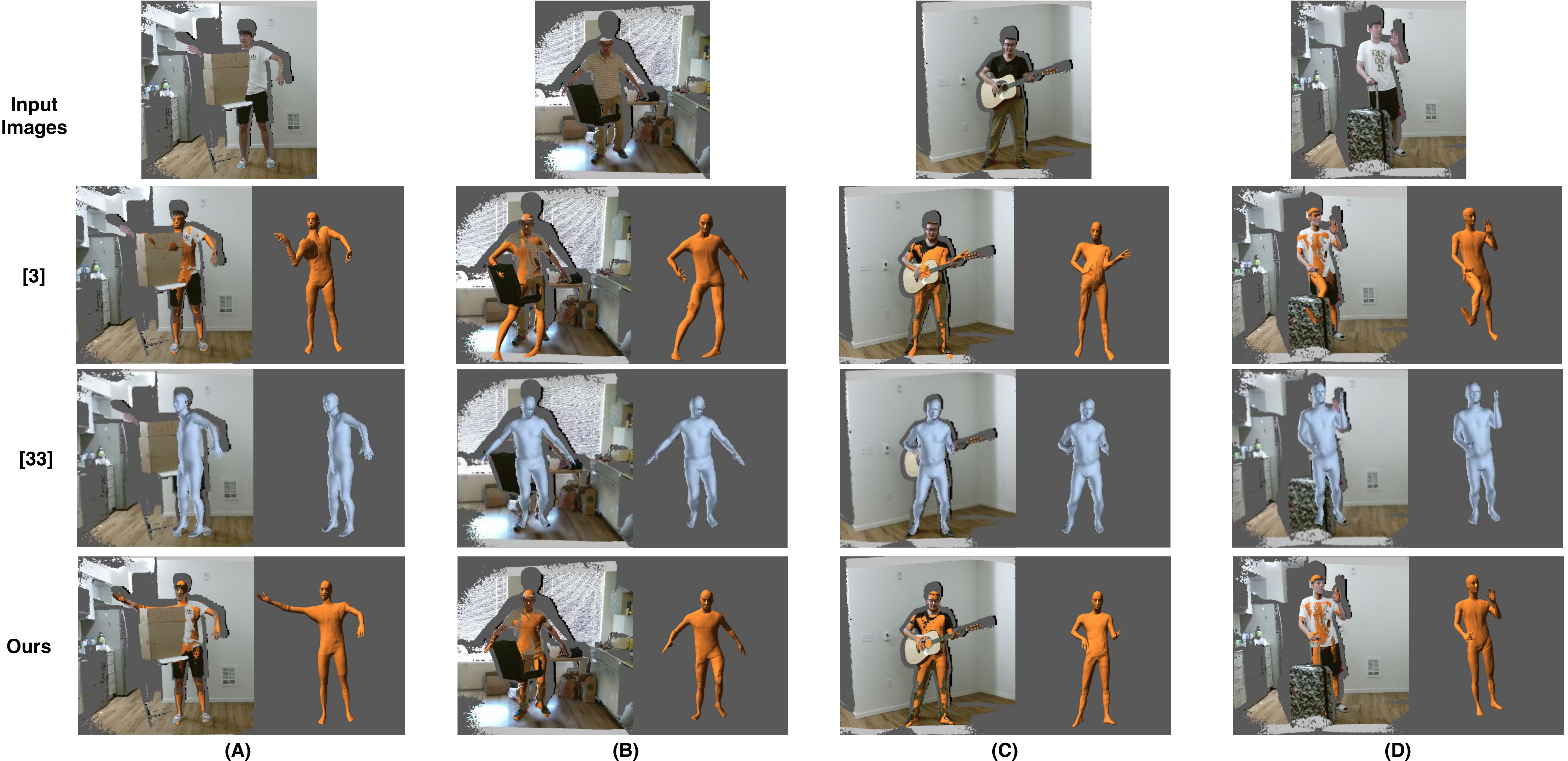}}
\caption{ \wwrevise{Our results compared against with \cite{walsman2017dynamic} and \cite{kanazawaHMR18}. Part (A) and (B) are tested with object-inserted video sequences. Part (C) and (D) are tested with human-object interaction video sequences. In each column the first row is the input video frame, the second row is failure cases using the geometric tracker alone \cite{walsman2017dynamic}, the third row is the results using direct regression approach \cite{kanazawaHMR18}, and the forth row is the results from our technique.} 
}
  \label{fig:result1}
\end{figure*}

Our primary task is accurate 3D tracking using RGB-D data, so we test our system using a metric which measures the
3D distance from ground truth joint positions.  We first test the performance of our hybrid tracker against
the method in \cite{walsman2017dynamic} on our augmented test set to examine the how much our Fast-FCN based filtering
improves overall tracking accuracy.  We then compare the Fast-FCN architecture against the larger U-Net \cite{ronneberger2015u} and VGG-FCN \cite{oliveira2016deep}
models and show that our simpler model performs almost as well as the U-Net and better than VGG-FCN in this context while being much faster.
We also report raw segmentation performance on this dataset for all three models.  Note that we do not report
performance on more popular segementation datasets such as the PASCAL part dataset \cite{chen2014detect}, because
these do not contain depth information, which is a critical component of our method. \wwrevise{
Additionally, in Figure \ref{fig:result1}, we provide test results of comparing our method against a geometric tracker \cite{walsman2017dynamic} and a state-of-art RGB based regression approach \cite{kanazawaHMR18}. The tests are performed using both object-inserted sequences and human-object interaction sequences from our dataset.}


All 
tests were performed on a machine running Ubuntu 16.04 with a 4GHz Intel i7 and an Nvidia GTX 1070 graphic card. 
We trained our Fast-FCN network using twenty-eight sequences that were augmented with additional objects.
A holdout set containing four video sequences with both segmentation and joint position ground truth,
and 2 object images were used for testing.  The human subjects and augmentation objects in this holdout set
were not featured in any of the training data in order to guarantee that we did not overfit to a particular
individual or occluding object.
\begin{figure}[thpb]
  \centering
  \hbox{
  \includegraphics[scale=0.33]{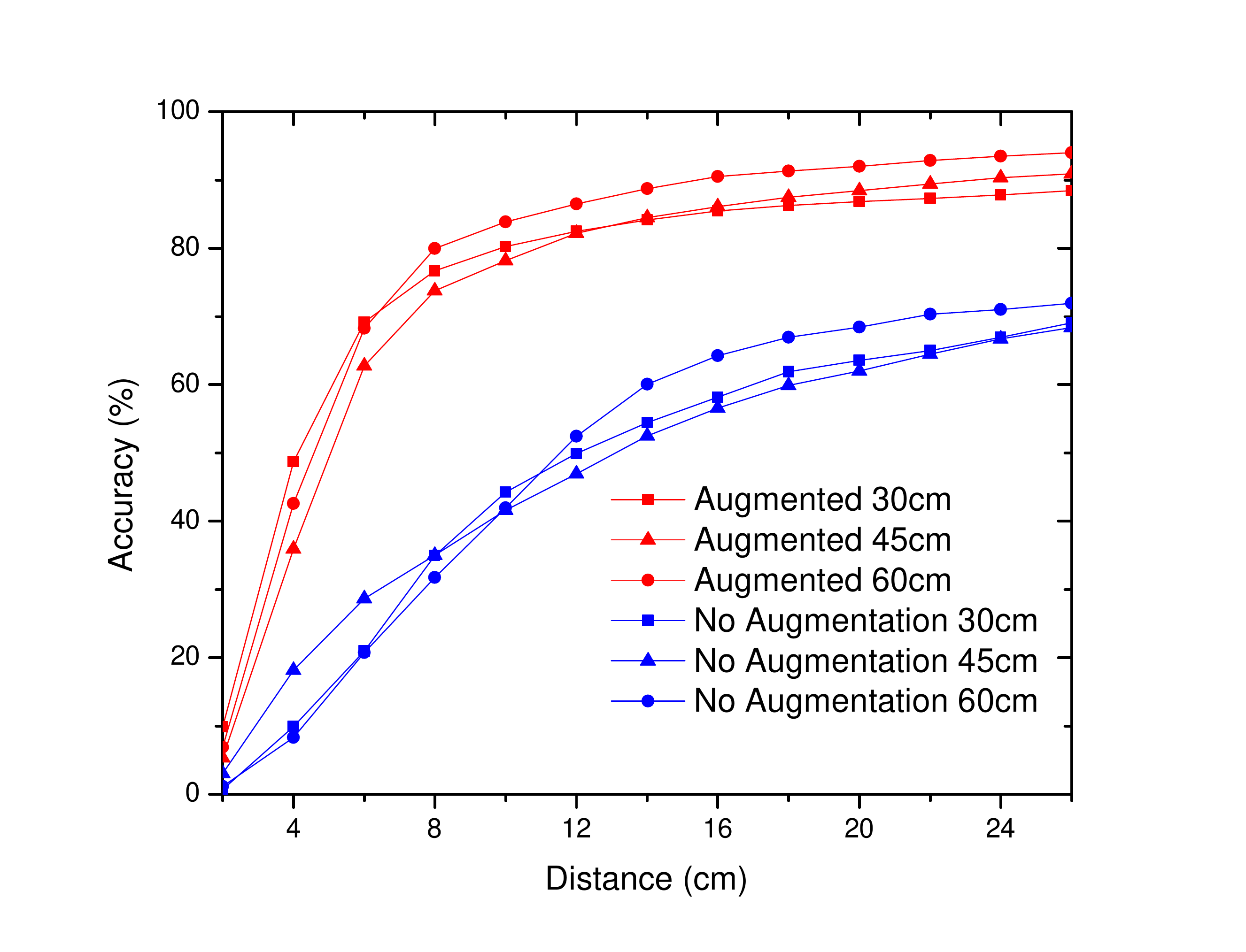}}
  \caption{Comparing our results with training by no occlusion dataset.
  }
  \label{fig:resultNonOcclusions}
\end{figure}

\subsection{Tracking accuracy}
We first compare our method with the tracking system proposed in \cite{walsman2017dynamic}. To simulate the scenarios with object interference, we inserted an object in front of the person slightly to the left or right, occluding around one third of the body in each video. Also, to test the system with different levels of interference, for the same sequence we separately inserted the object with the furthest point of the object at either 30cm, 45cm, 60cm away from the person. Figure \ref{fig:accuracy_plot} shows the accuracy of each method as a function of how close a joint must be to the ground-truth position in order to be considered accurate.  As the object is inserted closer to the person, the baseline method increasingly loses track and has a lower accuracy. However, with semantic segmentation, our system is able to perform with stability even in the most difficult scenarios.

\subsection{The Effects of Data Augmentation}
In order to evaluate the effect of data augmentation, we also trained a model on the original videos without adding
occluding objects.  The results are shown in \mbox{Figure \ref{fig:resultNonOcclusions}}.  Training on augmented data
is clearly beneficial.  It is also worth noting that the model trained without occlusions is actually worse
than running the tracker alone (compare to \mbox{Figure \ref{fig:accuracy_plot}}).  This shows that while good semantic
segmentation can help tracking performance, poor segmentation masks can be detrimental, because they can force data
association to bad observations.


\subsection{Architecture comparison}
We chose two state-of-art approaches in semantic segmentation with deconvolutional neural networks for comparison.
The first is U-Net \cite{ronneberger2015u}, which also inspired our Fast-FCN architecture.
In \cite{bulat2016human} \cite{xia2017joint} \cite{oliveira2016deep} \cite{nishi2017generation}, researchers developed FCN architectures by adding deconvolve layers based on the VGG-16 \cite{simonyan2014very} classification network. Thus, we chose the VGG-FCN structure proposed in \cite{oliveira2016deep} 
for body part labeling as our second comparison. 
We rebuilt the structure of these networks, and only modified the sizes to fit our dataset. We trained the networks on 
our dataset, and ran until convergence.
A summary of the structure of these three networks is shown in Table  \ref{table:structureTable}, along with the average
running time 
for a single frame input. \par
\begin{table}[b]
\centering
\caption{Number of each operation in FCN networks and average proceed time for an input}
\begin{tabular}{@{}ccccc@{}}
\toprule
Method  & Conv & Pool & Deconv & Processing Time (ms) \\ \midrule
VGG-FCN \cite{oliveira2016deep} & 15   & 5    & 5      & 12.21      \\
U-Net \cite{ronneberger2015u}   & 18   & 4    & 4      & 23.96      \\
Ours    & 9    & 3    & 3      & 7.86       \\ \bottomrule
\end{tabular}

\label{table:structureTable}
\end{table}
\begin{table}[thpb]
\centering
\caption{Pixel accuracy, background and non-background IoU for each network}
\begin{tabular}{@{}cccc@{}}
\toprule
Method  & Pix-Acc. & Background IoU & Non-background IoU \\ \midrule
VGG-FCN \cite{oliveira2016deep} & 97.34    & 93.26          & 54.55              \\
U-Net \cite{ronneberger2015u}   & 98.01    & 94.87          & 65.91              \\
Ours    & 97.65    & 94.19          & 62.38              \\ \bottomrule
\end{tabular}
\label{table:accurateTable1}
\end{table}

\begin{table}[thpb]
\centering
\caption{IoU of each body parts for each network}
\begin{tabular}{@{}cccclll@{}}
\toprule
Method  & \multicolumn{1}{l}{Head} & Torso & L-arm & R-arm & L-leg & R-leg \\ \midrule
VGG-FCN \cite{oliveira2016deep} & 75.72                    & 46.60 & 54.18 & 57.56 & 40.45 & 47.42 \\
U-Net \cite{ronneberger2015u}   & 79.14                    & 52.51 & 62.74 & 66.34 & 47.65 & 54.38 \\
Ours    & 78.16                    & 49.65 & 61.69 & 64.45 & 45.77 & 50.64 \\ \bottomrule
\end{tabular}
\label{table:accurateTable2}
\end{table}
\begin{figure}[thpb]
  \centering
  \hbox{
  \includegraphics[scale=0.33]{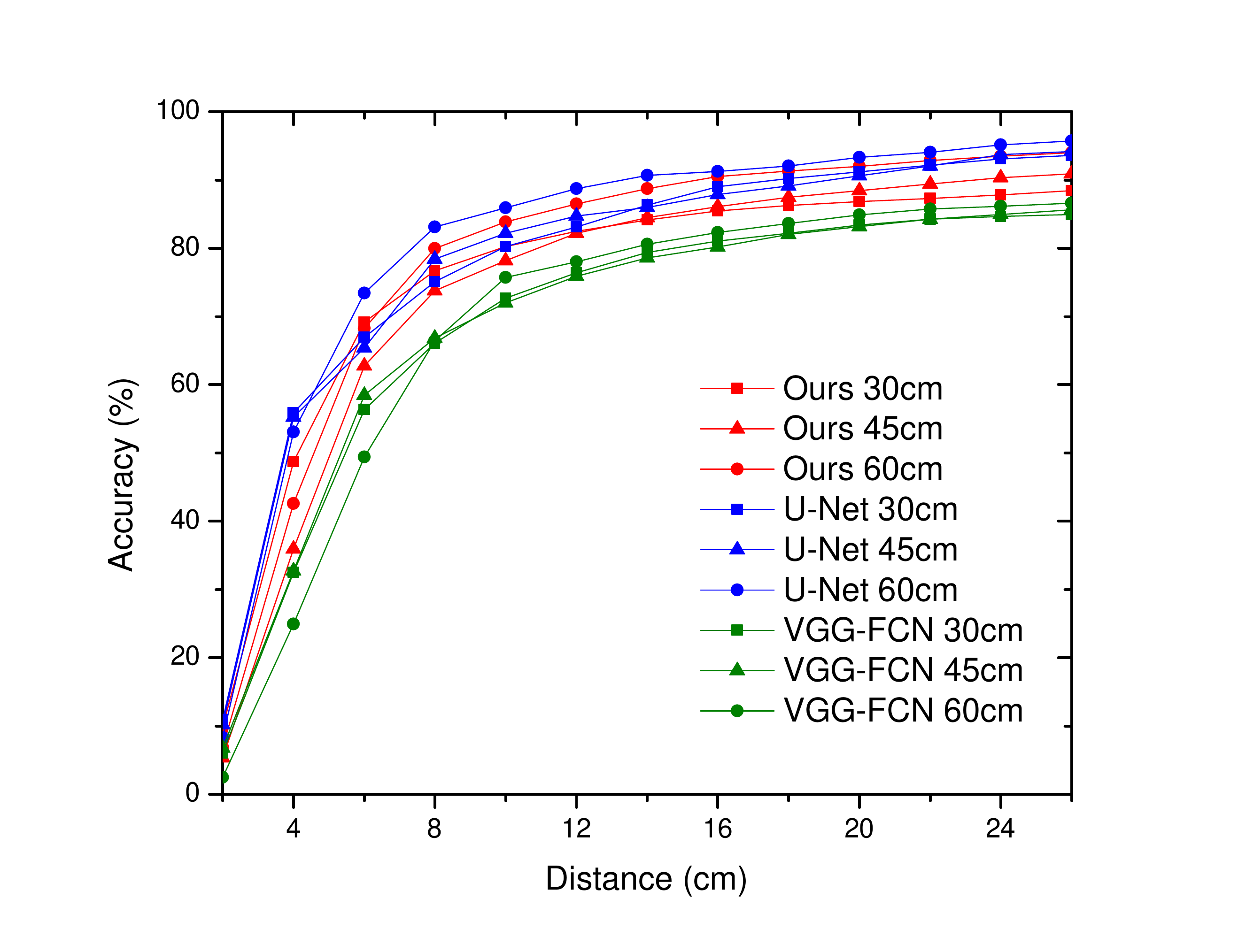}}
  \caption{The object-insertion test comparing with U-Net \cite{ronneberger2015u} and VGG-FCN \cite{oliveira2016deep} \atwnote{Should we combine this into one figure with the other one?}\wwnote{Well, I think they belongs to two separate subsection here. Also it might be good if we can show the improvement separately and conspicuously?}}
  \label{fig:architecture_plot}
\end{figure}
To illustrate the accuracy of each network on our dataset, we use pixel-wise accuracy and intersection over union (IoU) as evaluation metrics. The pixel-wise accuracy is calculated as the ratio of correctly labeled pixels over the total number of pixels. 
However, in our case, merely comparing pixel-wise accuracy is misleading because the background always occupies most of the area in the image, dominating the accuracy calculation. Thus, we also computed pixel-wise IoU for comparison, which is the
ratio
of the sum of correctly labeled pixels over the sum of
the union of 
similarly labeled pixels in the prediction and ground truth.\par

First, we test the networks with single frame images with objects inserted. The results in Table \ref{table:accurateTable1} and Table \ref{table:accurateTable2} show the accuracy of
the semantic segmentation output. Then we tested
tracking accuracy with the same video sequences and
data augmentation, and show the results in Figure \ref{fig:architecture_plot}.
We can see that our Fast-FCN model performs comparably to the U-Net, but at a third of the cost.

\section{Conclusion}

In this paper, we proposed a real-time solution for human pose tracking with occlusions based on fully convolutional neural networks and 3D articulated human model. The Fast-FCN we trained is able to accurately label the semantic segmentations of the person being tracked. By using the semantic segmentations, our system is able to initialize the starting pose automatically and track human poses in occlusions with high accuracy. We show that our model may be trained using a limited number of examples with no manual annotation.\par
 Many robotics applications such health care and personal assistance require very accurate 3D estimation of human pose. Our system furthers these goals by improving accuracy in challenging scenarios with clutter and occlusions.  We demonstrate the performance of our system 
 by comparing it with traditional geometric tracking, and the results show that our new method is able to significantly improve the tracking accuracy in the presence of human-object interactions.

\end{document}